\pdfoutput=1

\documentclass[11pt]{article}
\usepackage[table]{xcolor}

\usepackage[preprint]{acl}

\usepackage{times}
\usepackage{latexsym}
\usepackage{multirow}
\usepackage{booktabs,siunitx}
\usepackage{enumitem}
\usepackage{array}
\usepackage{moresize}
\usepackage{stfloats}
\usepackage{color}
\usepackage{pifont}
\usepackage{amssymb}
\usepackage{float}
\usepackage{soul}

\newcommand{\cmark}{\ding{51}}%
\newcommand{\xmark}{\ding{55}}%

\definecolor{lightgreen}{rgb}{.7,.9,.1}

\usepackage{paracol}
\usepackage[many]{tcolorbox}
\usepackage{lipsum}
\NewTColorBox{NewBox}{ s O{!htbp} }{%
  floatplacement={#2},
  IfBooleanTF={#1}{float*,width=\textwidth}{float},
  colframe=blue!50!black,colback=blue!10!white,
  }

\usepackage{xcolor}

\usepackage{tablefootnote}

\usepackage[T1]{fontenc}

\usepackage[utf8]{inputenc}

\usepackage{microtype}

\usepackage{inconsolata}

\title{LionGuard: Building a Contextualized Moderation Classifier to Tackle Localized Unsafe Content}

\author{Jessica Foo\thanks{Equal contribution}\\
  GovTech Singapore \\
  \texttt{jessica\_foo@tech.gov.sg} \\\And
  Shaun Khoo\footnotemark[1]\\
  GovTech Singapore \\
  \texttt{shaun\_khoo@tech.gov.sg} \\}

\graphicspath{{./images/}}

\begin{document}
\maketitle
\begin{abstract}

As large language models (LLMs) become increasingly prevalent in a wide variety of applications, concerns about the safety of their outputs have become more significant. Most efforts at safety-tuning or moderation today take on a predominantly Western-centric view of safety, especially for toxic, hateful, or violent speech. In this paper, we describe LionGuard, a Singapore-contextualized moderation classifier that can serve as guardrails against unsafe LLM outputs. When assessed on Singlish data, LionGuard outperforms existing widely-used moderation APIs, which are not finetuned for the Singapore context, by 14\% (binary) and up to 51\% (multi-label). Our work highlights the benefits of localization for moderation classifiers and presents a practical and scalable approach for low-resource languages.\\
\textcolor{red}{Warning: this paper contains references and data that may be offensive.}
\end{abstract}

\section{Introduction}
Large language models (LLMs) have led to a breakthrough in the generative abilities of conversational AI agents, achieving unprecedented levels of linguistic fluency and generalizability. Due to their strong conversational abilities, LLMs have been deployed to a range of domains such as workplace productivity, education, and customer service. 

Given how frequently and directly users interact with these systems, moderation guardrails have been proposed to safeguard against the risk of LLMs generating harmful content. This is especially crucial for LLMs which may not have been sufficiently safety-tuned, as they can be easily instructed to generate hateful, toxic and offensive material at scale. In addition, moderation classifiers can also be used to automatically generate adversarial data \citep{perez} for safety alignment in LLM model training, score outputs from red-teaming, and benchmark LLMs in unsafe content generation. Hence, a robust and accurate moderation classifier can help in building safe LLMs at both the output (as guardrails) and input (augmenting training data) level. 

The most widely used content moderation classifiers today include 
OpenAI's Moderation API\footnote{\url{ https://platform.openai.com/docs/guides/moderation/overview}}, Jigsaw's Perspective API,\footnote{\url{ https://developers.perspectiveapi.com/s/about-the-api-attributes-and-languages?language=en_US}} and Meta's LlamaGuard.\footnote{\url{https://huggingface.co/meta-llama/LlamaGuard-7b}} While these classifiers are continually updated and have gradually incorporated multilingual capabilities \citep{perspective-multilingual}, they have not been tested rigorously on low-resource languages. Singlish, an English creole (i.e. a variant of English) is widely used by people residing in Singapore, with a population of close to 5.5 million. As a creole language, Singlish has acquired its own unique phonology, lexicon and syntax \citep{singlish}. As such, the linguistic shift between English and Singlish is significant enough such that existing moderation classifiers that perform well on English are unlikely to perform well on Singlish. 

We present a practical and scalable approach to localizing moderation, which can be applied to any low-resource language. In this work, we make the following contributions:

\begin{itemize}[nolistsep]
  \item \emph{Defining a safety risk taxonomy aligned to the local context.} We constructed our safety risk taxonomy by combining existing taxonomies across various commercial providers and aligning these categories with relevant Singaporean legislation and guidelines, such as the Singapore Code of Internet Practice.\footnote{\href{https://www.imda.gov.sg/-/media/imda/files/regulations-and-licensing/regulations/codes-of-practice/codes-of-practice-media/policiesandcontentguidelinesinternetinternecodeofpractice.pdf}{IMDA's Singapore Code of Internet Practice}}
  \item \emph{Creating a new large-scale dataset of Singlish texts for training moderation classifiers.} We collected Singlish texts from various online forums, conducted automated labelling using safety-tuned LLMs\footnote{We used OpenAI's GPT-3.5-Turbo (version 0613), Anthropic's Claude 2.0, and Google's PaLM 2 (text-bison-002).} using our safety risk taxonomy, resulting in a novel dataset of 138k Singlish texts that can be used for safety-tuning or benchmarking LLMs, or developing moderation classifiers.
  \item \emph{Contextualized moderation classifier outperforms generalist classifiers.} We finetuned a range of classification models on our automatically labelled dataset, and our best performing models outperformed Moderation API, Perspective API and LlamaGuard, while being faster and cheaper to run than using safety-tuned LLMs as guardrails. LionGuard is available on Hugging Face Hub.\footnote{\url{ https://huggingface.co/govtech/lionguard-v1}}
\end{itemize}

\section{Singlish, an English Creole}
Singlish is mainly influenced by non-English languages like Chinese, Malay, Tamil and Chinese dialects (e.g., Hokkien). While based on English, different languages are often combined within single utterances. To illustrate with the example below, the phrase "\textit{chionging}" is derived from the Chinese romanized word "\textit{chong}", which means "to rush"; the "\textit{-ing}" indicates the progressive verb tense from English grammar; "\textit{lao}" is the romanization of the Chinese word that means "old"; "\textit{liao}" is a Singlish particle that means "already".

\begin{tcolorbox}[width=\linewidth, sharp corners=all, colback=white!95!black]
\textit{"Either they just finished their shift work, having their supper after chionging or the lao uncles who are drinking there for a few hours liao."} (Comment from HardwareZone, posted on Sep 2023)
\end{tcolorbox}

Singlish also contains content-specific terminology. For example, "\textit{ceca}", the Singlish racial slur which describes people of Indian nationality, is a derogatory synecdoche. It refers to the Comprehensive Economic Cooperation Agreement (CECA), a free-trade agreement signed between Singapore and India which has faced scrutiny in recent years.\footnote{\url{https://str.sg/3J4U}} Furthermore, new vocabulary has emerged in the online domain, such as the word "\textit{sinkie}", which is a self-derogatory term referring to Singaporeans. Such lexicons are unlikely to be understood by Western-centric language models, unless they have been specifically trained or instructed to.

\begin{tcolorbox}[width=\linewidth, sharp corners=all, colback=white!95!black]
\textit{"you sinkie? if sinkie is good thing. if ceca, best to diam diam ah if not tio hoot by sinkies."} (Comment from HardwareZone, posted on Aug 2020)
\end{tcolorbox}

Several works have emerged to tackle Singlish for various Natural Language Processing (NLP) tasks, including sentiment analysis \citep{singlish-sentiment-1, singlish-sentiment-2, singlish-sentiment-3}, parts-of-speech tagging \citep{singlish-pos} and neural machine translation \citep{singlish-nmt}. \citet{hsieh} trained a Singlish BERT model to identify Singlish sentences, while \citet{singbert_embedding} fine-tuned BERT on a colloquial Singlish and Manglish\footnote{Informal form of Malaysian English} corpus (SingBERT). Such efforts highlight the significant linguistic differences between English and Singlish and the need for Singlish-focused content moderation.

\section{Related Work}

\subsection{Content moderation}

The importance of content moderation has led to a plethora of works focused on the detection of toxic and abusive content \citep{Nobata2016AbusiveLD, de-gibert-etal-2018-hate, chakravartula-2019-hateminer, berttransfer, vidgen}. Bidirectional Encoder Representations from Transformers (BERTs) \citep{bert} first emerged as powerful word embeddings that could be fine-tuned for downstream tasks like hate speech detection. \citet{Lee_2021} used BERT embeddings in the Hateful Memes Challenge \citep{hatefulmemes}, while \citet{liu-etal-2019-nuli} combined BERT and a Long-Short Term Memory model on the OffensEval dataset \citep{zampieri-etal-2019-semeval}. \citet{hatebert_embedding} re-trained BERT on offensive Reddit comments, building a shifted BERT model, HateBERT, that outperformed general BERT in hate speech detection.  

Moderation APIs have become increasingly popular due to the ease at which they can be integrated into applications. Such APIs aim to be universally applicable to different languages and domains. \citet{perspective} developed Perspective API, which uses multilingual BERT-based models that are then distilled into single-language Convolutional Neural Networks (CNNs) for each language supported. \citet{openaimoderation} developed OpenAI's Moderation API, which uses a lightweight transformer decoder model with a multi-layer perceptron head for each toxicity category. However, one concern amidst the increasing adoption of moderation APIs is how strikingly different toxicity triggers are across the Western and Eastern contexts \citep{Chong2022UnderstandingTT}, underscoring the importance of localized content moderation.


\subsection{Low-resource language adaptation for moderation}

Adapting toxicity detection to Singlish, \citet{yunting-hatespeech} used a CNN to detect hate speech from Twitter data. \citet{haber-etal-2023-improving} curated a multilingual dataset of Reddit comments in Singlish, Malay and Indonesian and found that domain adaption of mBERT \citep{bert} and XLM-R \citep{xlmr} models improved F1 performance in detecting toxic comments. \citet{totaldefmeme} analyzed multimodal Singlish hate speech by creating a dataset of offensive memes. Our work contributes to this space by establishing a more systematic approach to detecting unsafe content with automated labelling and by developing a contextualized moderation classifier which outperforms existing generalized moderation APIs.


\subsection{Automated labelling}
Despite requiring more time and resources, human labelling has frequently been used to generate gold standard labels for toxic speech, particularly via crowdsourcing \citep{Davidson_Warmsley_Macy_Weber_2017, parrish-etal-2022-bbq}. However, \citet{waseem-2016-racist} found that amateur annotators were more likely than expert annotators to label items as hate speech, causing poorer data quality. Considering the scale of data required for building safe LLMs, automated labelling has emerged as an alternative to human labelling. \citet{bai2022constitutional} used Constitutional AI to automatically perform evaluations with Claude, and then trained a preference model using the dataset of AI preferences. \citet{chiuhatespeech} found that with few-shot learning, GPT-3 can be used to detect sexist or racist text. \citet{plaza-del-arco-etal-2023-respectful} also found that zero-shot prompting of FLAN-T5 produced favorable results on several hate speech benchmarks. \citet{llamaguard} proposed an LLM-based input-output safeguard model, LlamaGuard, which classifies text inputs based on specific safety risks as defined by prompts. Unlike existing works that rely on a single model for automated labelling, we combined several LLMs to provide more accurate and reliable labels, leveraging the collective wisdom and knowledge of several safety-tuned LLMs.

\section{Methodology}

\begin{figure*}
\centering
    \includegraphics[width=2.1\columnwidth]{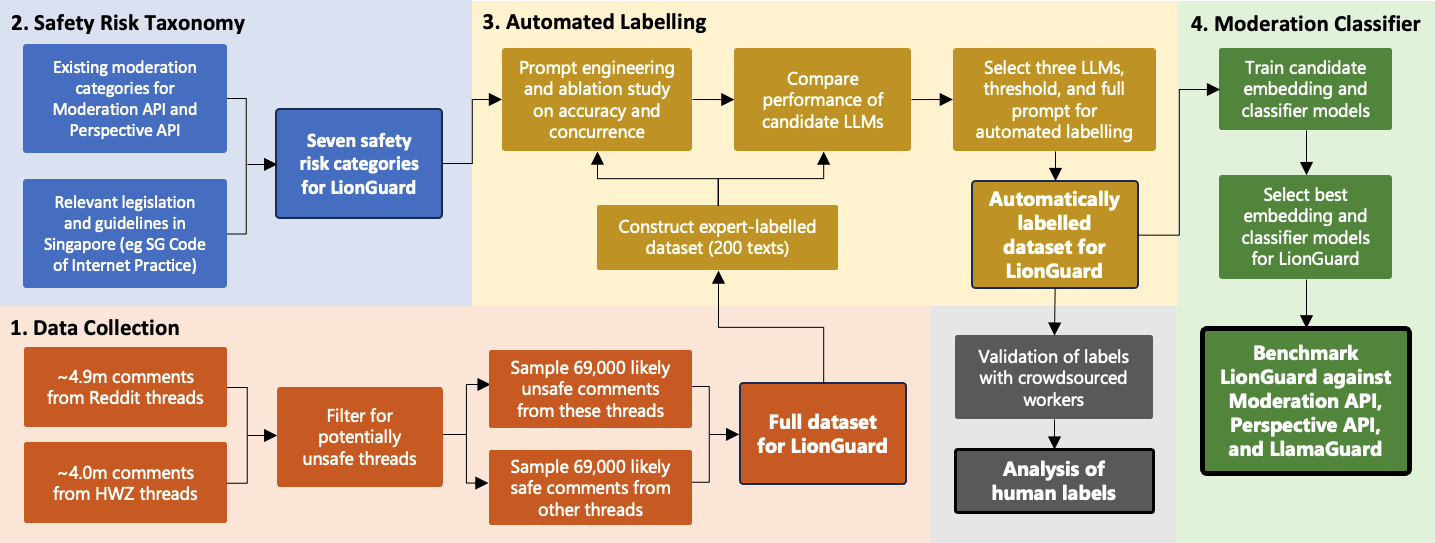}
\caption{Overview of the 4-step methodology in building LionGuard}
\label{fig:methodology}
\end{figure*}

To develop a robust moderation classifier that is sensitive to Singlish and Singapore's context, we adopted a 4-step methodology as seen in Figure \ref{fig:methodology}.

\subsection{Data Collection}
\label{sec:methodology-data}

To build a dataset of Singlish texts, we collected comments from HardwareZone's Eat-Drink-Man-Woman online forum and selected subreddits from Reddit on Singapore.\footnote{r/Singapore, r/SingaporeHappenings, r/SingaporeRaw} The former is notorious in Singapore as a hotspot of misogynistic, xenophobic, and toxic comments,\footnote{\url{https://www.ricemedia.co/pretty-privilege-bbfa/}} while the latter is a popular online forum for Singapore-specific issues. We collected comments on all threads between 2020 and 2023 from both forums, resulting in a dataset of approximately 8.9 million comments.


However, upon manual inspection of the data, only a small minority of the comments were unsafe. Both forums have a wide range of topics which are not always controversial or harmful, and forum moderators often remove the most toxic comments.

To ensure sufficient unsafe texts for a balanced dataset, we used entire threads that discussed controversial topics in Singapore or contained offensive words (see Appendix \ref{sec:controversial-topics}), which were more likely to be unsafe. This resulted in approximately 400,000 texts, which we randomly subsampled to a smaller set of 69,000 potentially unsafe texts. We then randomly sampled another 69,000 texts from the remaining dataset that had not been identified as likely to be unsafe, for greater heterogeneity in topics and language, resulting in a final training dataset of 138,000 texts (examples in Appendix \ref{sec:singlish-examples}).

\subsection{Safety Risk Taxonomy}
\label{sec:methodology-categories}

Next, we referenced the moderation categories defined in OpenAI's Moderation API, Jigsaw's Perspective API and Meta's LlamaGuard, and took into consideration Singapore's Code of Internet Practice and Code of Practice for Online Safety.\footnote{\url{https://www.imda.gov.sg/-/media/imda/files/regulations-and-licensing/regulations/codes-of-practice/codes-of-practice-media/code-of-practice-for-online-safety.pdf}} 


We defined seven categories of safety risks for LionGuard. Brief descriptions of each category are listed below, while the full definitions are available in Appendix~\ref{sec:app-toxicity-categories}. The key differences between our safety risk categories and OpenAI's, Jigsaw's and Meta's are summarized in Table \ref{table:toxicity-categories}.

\begin{enumerate}[nolistsep]
    \item \textbf{Hateful}: Content that expresses, incites, or promotes hate based on race, gender, ethnicity, religion, nationality, sexual orientation, disability status, or caste. 
    \item \textbf{Harassment}: Content that expresses, incites, or promotes harassing language towards any target/individual.
    \item \textbf{Encouraging public harm}: Content that promotes, facilitates, or encourages harmful public acts, vice or organized crime.
    \item \textbf{Encouraging self-harm}: Content that promotes or depicts acts of self-harm, such as suicide, cutting, and eating disorders. 
    \item \textbf{Sexual}: Content meant to arouse sexual excitement, such as the description of sexual activity, or that promotes sexual services (excluding sex education and wellness). 
    \item \textbf{Toxic}: Content that is rude, disrespectful, or profane, including the use of slurs. 
    \item \textbf{Violent}: Content that depicts death, violence, or physical injury. 
\end{enumerate}

\begin{table}[h!]
\centering
 \begin{tabular}{m{1.5cm} m{1.5cm} m{1.5cm} m{1.5cm}} 
 \hline 
 \textbf{LionGuard} & \textbf{OpenAI} & \textbf{Jigsaw} & \textbf{LlamaGuard} \\ [0.5ex] 
 \hline
 Hateful & Hate & Identity\newline attack & Violence and Hate\\
 \hline
 Harassment & Harassment & Insult & -\\
 \hline
 Public harm & - & - & Crime\tablefootnote{LlamaGuard defines separate categories for Guns and Illegal Weapons, Regulated or Controlled Substances, Criminal Planning.}\\
 \hline
 Self-harm & Self-harm & - & Self Harm\\
 \hline
 Sexual & Sexual & - & Sexual \\
 \hline
 Toxic & - & Toxicity,\newline Profanity & -\\
 \hline
 Violent & Violence & Threat & Violence and Hate\\
 \hline
 \end{tabular}
\caption{High-level comparison of content moderation categories across LionGuard, OpenAI's Moderation API, and Jigsaw's Perspective API.}
\label{table:toxicity-categories}
\end{table}

\subsection{Automated Labelling}
\label{sec:methodology-llmlabel}

We then automatically labelled our Singlish dataset according to our safety risk categories using LLMs. Automated labelling with LLMs is increasingly popular given vast improvements in instruction-following with recent LLMs \citep{openai-instruction, bloomz, wizardlm}. 

To verify the accuracy of our automated labelling, we internally labelled 200 texts that served as our expert-labelled dataset. The dataset was handpicked by our team with a focus on selecting particularly challenging texts that were likely to be mislabelled. This consisted of 143 unsafe texts (71.5\%) and 57 safe texts (28.5\%). 

\subsubsection{Engineering the labelling prompt}
\label{sec:methodology-llmlabel-prompteng}

We incorporated the following prompt engineering methods for our automated labelling:

\begin{enumerate}[nolistsep]
    \item \textbf{Context prompting} \citep{openai-details-prompting}: We specified that the text to be evaluated is in Singlish and that the evaluation needs to consider Singapore's socio-cultural context. We also provided examples and definitions of common offensive Singlish slang.
    \item \textbf{Few-shot prompting} \citep{few-shot-prompting}: We gave examples of Singlish texts (that included Singlish slang and Singaporean references) and associated safety risk labels.
    \item \textbf{Chain-of-Thought (CoT) prompting} \citep{chain-of-thought-prompting}: We specified each step that the LLM should take in evaluating the text, asking it to consider whether the text fulfils any of the seven criteria, and to provide a "yes/no" label along with a reason for its decision.
\end{enumerate}

To determine the effectiveness of these prompt engineering techniques, we conducted an ablation study that compared the performance of the:
\begin{enumerate}[label=(\alph*),nolistsep]
    \item Full prompt (combining all three methods)
    \item Full prompt \textit{less} context prompting (no Singlish examples)
    \item Full prompt \textit{less} few-shot prompting
    \item Full prompt \textit{less} CoT prompting
\end{enumerate}

We measured how effective the prompts were in terms of their F1 score (i.e. taking into account precision and recall of detecting unsafe content with respect to our expert-labelled dataset)\footnote{Note that F1-score is measured using only texts which there was a consensus across all LLMs on whether the text was safe or unsafe. This is because we trained our moderation classifier only on texts where there was a consensus to preserve higher quality in the training dataset. This is explained in subsection \ref{sec:determining-threshold}.} and agreement (i.e. how frequently the LLMs concurred). F1 scores were chosen as our evaluation metric as the expert-labelled dataset was slightly skewed to unsafe text (71.5\%).

\begin{figure}[h]
  \centering
  \includegraphics[width=0.8\columnwidth]{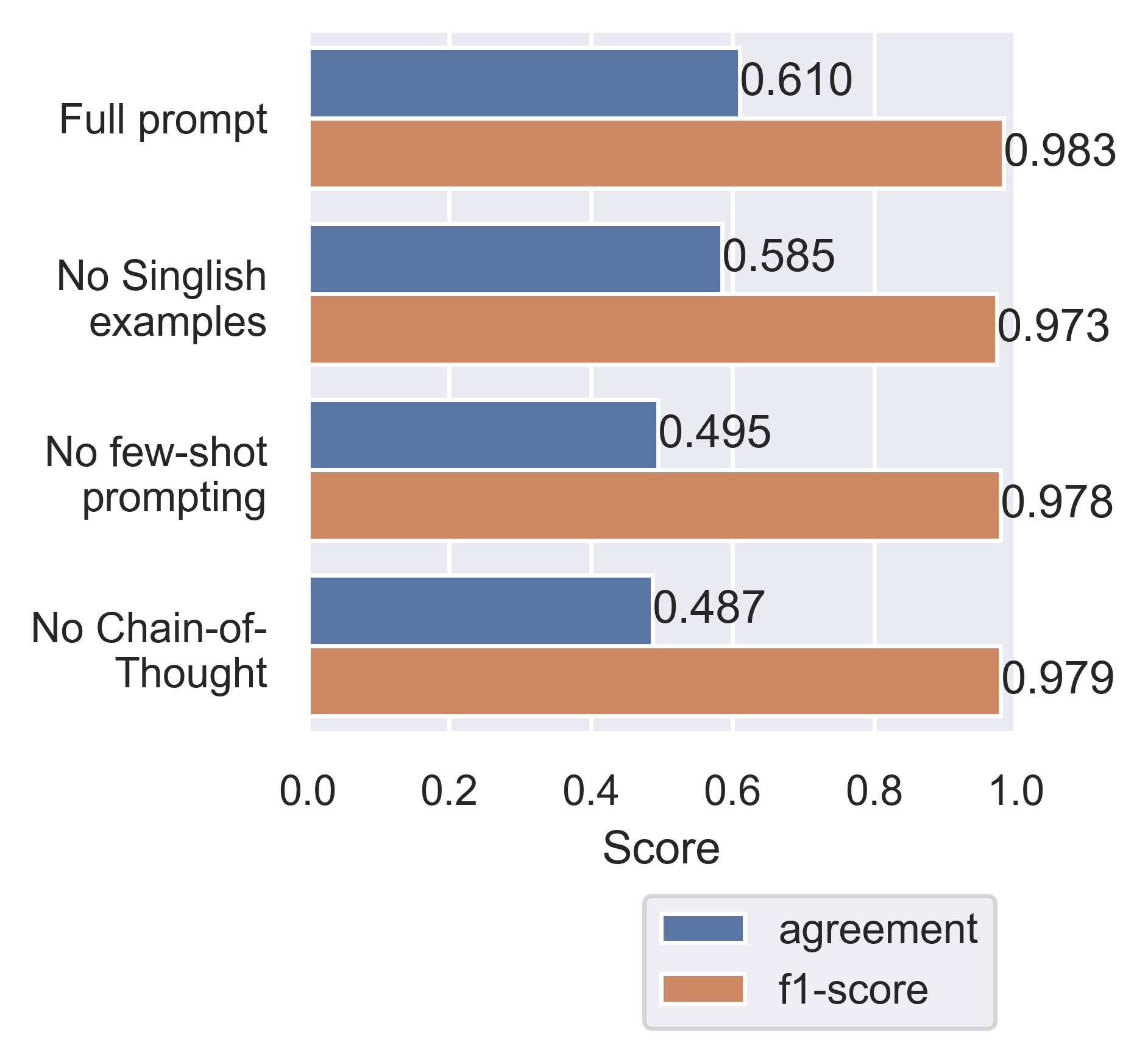} 
  \caption{F1 scores and agreement across the 4 candidate LLMs for the prompt ablation comparison}
  \label{fig:prompt-ablation}
\end{figure}

As seen in Figure \ref{fig:prompt-ablation}, we found that using all three approaches together was most effective in terms of both F1 scores and agreement.

\subsubsection{LLM Selection}
\label{sec:methodology-llmlabel-llmselect}

We started with four candidate LLMs: OpenAI's GPT-3.5-turbo (version 0613) \citep{openai-chatgpt-3.5-turbo}, Anthropic's Claude 2.0 \citep{anthropic-claude2}, Google's PaLM 2 (text-bison-002) \citep{google-palm2}, and Meta's Llama 2 Chat 70b \citep{meta-llama-2}. These LLMs were chosen as they were the top-performing LLMs at the time and had also been safety-tuned. 

We assessed each LLM's accuracy by comparing their F1 scores in labelling texts against the expert-labelled dataset. We ran all four prompts detailed in subsection \ref{sec:methodology-llmlabel-prompteng} for each of the candidate LLMs.\footnote{We were unable to get a valid label from Llama 2 for one Reddit text using the prompt template without CoT, despite varying temperature and top\_p parameters. We chose to drop it from the analysis, so all scores reported for Llama 2 for the prompt without CoT are with 199 texts instead of the full 200 texts. This does not change the results significantly since we ultimately chose the full prompt approach.}

\begin{figure}[h]
  \centering
  \includegraphics[width=0.8\columnwidth]{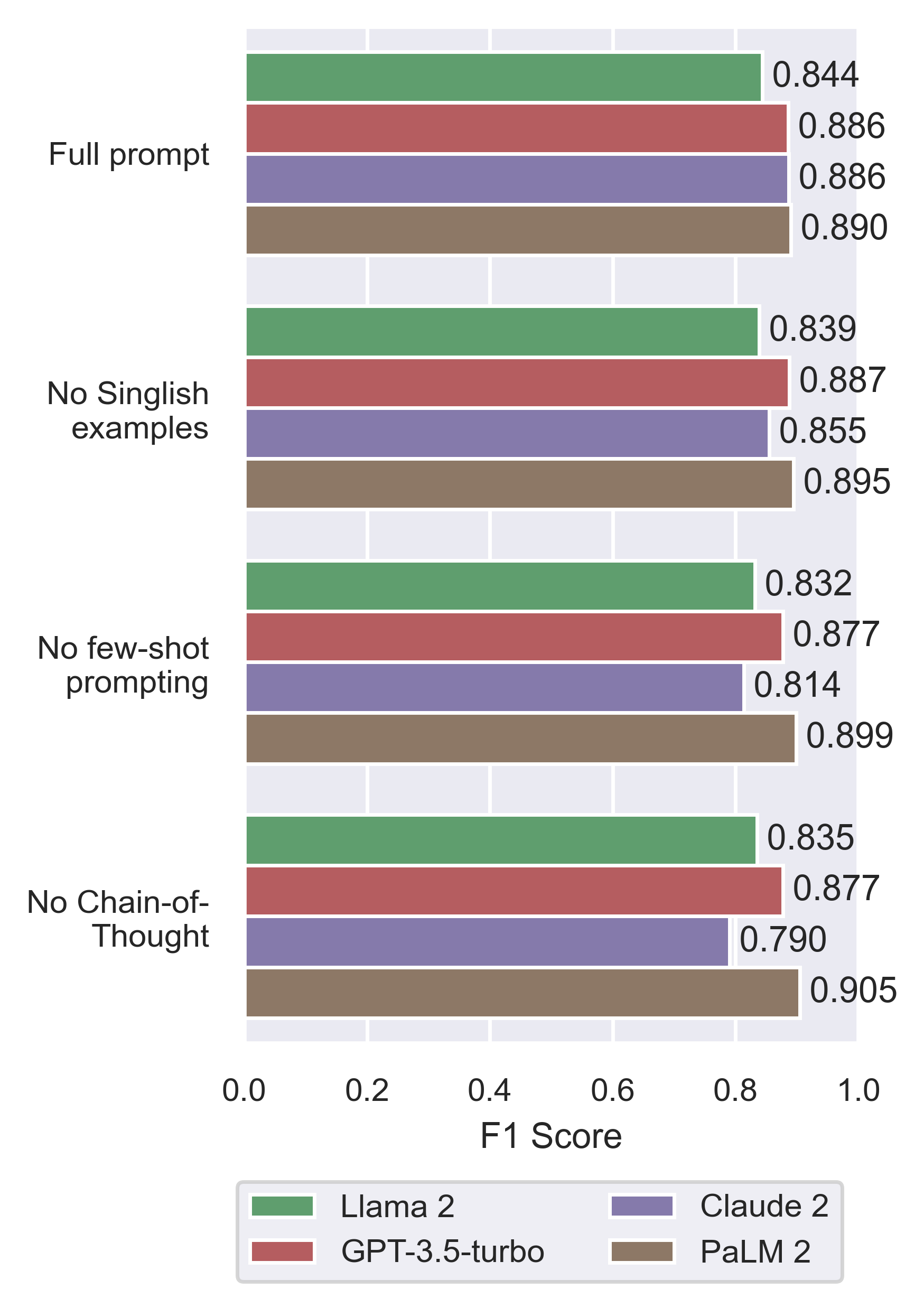} 
  \caption{F1 scores for each combination of prompt and candidate LLM}
  \label{fig:candidate-llm-accuracy}
\end{figure}



As seen in Figure \ref{fig:candidate-llm-accuracy}, Llama 2 underperformed by a clear margin compared to the other three candidate LLMs when the full prompt was used. Upon closer inspection, we found that Llama 2 predicted nearly every text as unsafe,\footnote{Because of this, Llama 2 had a recall of 1 and a precision of 0.730. This is in contrast to the other LLMs which had significantly higher precision scores of 0.830 (GPT-3.5-turbo), 0.967 (Claude 2), and 0.826 (Palm 2).} and this behaviour persisted despite additional changes to the prompt. Through error analysis (see Appendix \ref{sec:llama2-error-analysis}), we found that Llama 2 was overly conservative and provided incorrect justifications for classifying safe text as unsafe. As such, Llama 2 was dropped to avoid distorting the labels for our classification dataset.

\subsubsection{Determining the Threshold for Safety}
\label{sec:determining-threshold}

\begin{figure}[h]
  \centering
  \includegraphics[width=0.8\columnwidth]{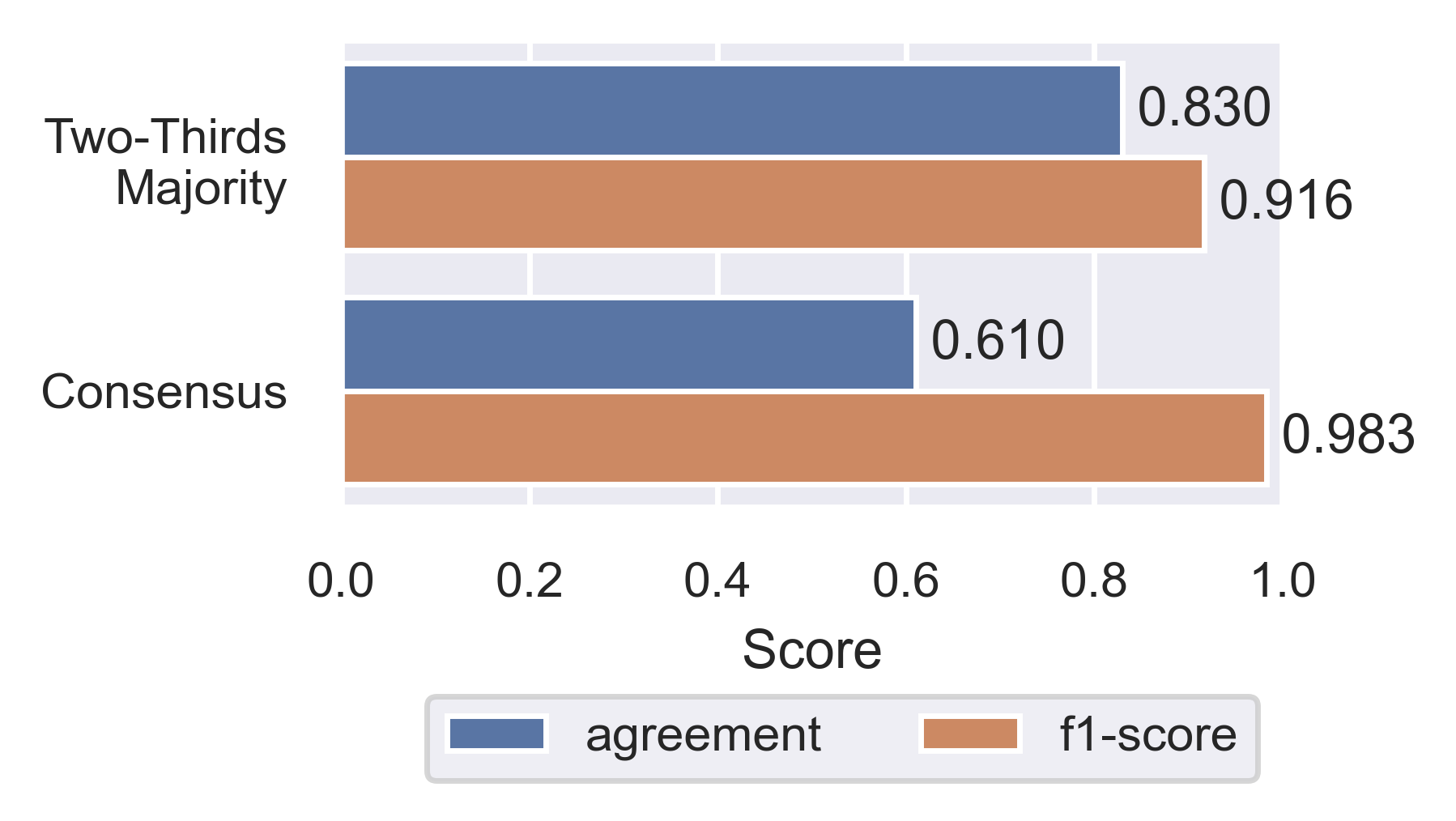} 
  \caption{Comparing F1 scores and agreement for different threshold levels}
  \label{fig:threshold-levels}
\end{figure}

After determining the best prompt and set of LLMs for labelling, we considered two thresholds for determining unsafe content: majority vote (i.e. at least two of three LLMs label the text as unsafe) or consensus (i.e. all 3 LLMs label the text as unsafe). We compared the F1 scores and agreement for these two threshold levels, as seen in Figure \ref{fig:threshold-levels}.

As we were assembling a new dataset to build a contextualized moderation classifier from scratch, we determined that the priority was labelling accuracy. As such, we chose the consensus approach for our training (see subsection \ref{sec:classifier-training}), which had higher accuracy even though the agreement rate is lower. 

\subsubsection{Compiling the dataset}
\label{sec:compiling-dataset}

The final dataset consisted of 138,000 labelled texts. The breakdown of the number of positive labels in the dataset can be found in Table \ref{table:category-counts}. Note the severe imbalance of data for most categories, which made our model training process challenging. The dataset was split into train (70\%), validation (15\%), and test (15\%) sets. Texts from the same threads were allocated to the same split. All experimental results in section \ref{sec:classifier-results} are reported using the test set.

\begin{table}[h!]
\centering
 \begin{tabular}{m{2.5cm} m{2.5cm}} 
 \hline
 \textbf{Category} & \textbf{Positive labels}\\ [0.5ex] 
 \hline
 \texttt{hateful} & 537 (0.40\%)\\
 \texttt{harassment} & 101 (0.07\%)\\
 \texttt{public harm} & 147 (0.11\%)\\
 \texttt{self-harm} & 82 (0.06\%) \\
 \texttt{sexual} & 695 (0.51\%) \\
 \texttt{toxic} & 7,295 (7.30\%)\\
 \texttt{violent} & 153 (0.11\%)\\
 \hline\hline
\texttt{unsafe} & 8,375 (6.15\%)\\
 \end{tabular}
\caption{Breakdown of the number of positive labels in the dataset. Note that the sum of all seven categories do not equal to the number of positive binary labels (\texttt{unsafe}) as a text can satisfy more than one category.}
\label{table:category-counts}
\end{table}

\subsection{Moderation Classifier}
\label{sec:classifier-training}


\textbf{Architecture}: LionGuard, our moderation classifier, comprises two components: an embedding and classifier model. The embedding model generates a vector representation of the text, which the classifier model uses as inputs to generate a moderation score. This simple architecture enables us to test different embedding and classifier models to find the best-performing combination for LionGuard.

\textbf{Embedding model}: Our approach compared general embedding models against finetuned models. We chose BAAI General Embedding (BGE) \citep{bge_embedding} given its strong performance on Hugging Face's leaderboard for embeddings,\footnote{\url{https://huggingface.co/spaces/mteb/leaderboard}} HateBERT \citep{hatebert_embedding}, as well as SingBERT \citep{singbert_embedding}. We also experimented with masked language modelling (MLM) on these embedding models on a separate sample of 500,000 texts from our initial dataset of 8.9m texts for 30 epochs. Ablation studies were also conducted with BGE-small, BERT-base and BERT-large embedding models.


\textbf{Classifier model}: We selected our classifier models based on different levels of model complexity to reveal any differences in performance due to the number of parameters. In order of complexity, we chose a ridge regression classifier, XGBoost classifier, and a neural network (consisting of one hidden and one dropout layer). We carried out hyperparameter tuning for the XGBoost and neural network classifier. More details on the hyperparameter search and the final set of hyperparameters are provided in Appendix \ref{sec:hyperparameter-results}.


\textbf{Training}: We developed two versions of LionGuard: a binary classifier (to detect if a text is \texttt{safe} or \texttt{unsafe}), and a multi-label classifier (to detect if a text fulfills any category in our safety risk taxonomy defined in \ref{sec:methodology-categories}). 


For the binary classifier, we limited the training data to texts where there was consensus among the LLMs on the label (i.e., \texttt{unsafe} or \texttt{safe}). This resulted in a smaller dataset of 99,597 texts (72.2\%).

For the multi-label classifier, we trained a dedicated classifier model for each category. We included texts where there was a consensus for that category, which enabled us to maximize the use of our limited number of positive labels. Apart from the \texttt{toxic} category, there was consensus on over 96\% of the labels for each of the other categories.\footnote{For the \texttt{toxic} category, the consensus rate was 72.4\%. Although this meant there was less training data for the \texttt{toxic}-specific classifier, there was still more than enough training data (around 99,900 texts). Moreover, the \texttt{toxic} category also had more positive labels than the other categories.} 

\textbf{Evaluation}: Due to the heavily imbalanced dataset, we chose the Precision-Recall AUC (PR-AUC) as our evaluation metric as it can better represent the classifier's ability to detect unsafe content across all score thresholds. PR-AUC was also used by OpenAI \citep{openaimoderation} and LlamaGuard \citep{llamaguard} in their evaluations. 

\textbf{Benchmarking}: After identifying the best classifier, we compared LionGuard with Moderation API, Perspective API and LlamaGuard. Both APIs provided a score, while LlamaGuard provided the probability of the first (classification) token, which we used to calculate the PR-AUC. We benchmarked them on both the binary and multi-label experiments. 

\section{Results}

\subsection{Validation of LLM labels with humans}

We compared labels provided by the LLMs with labels annotated by crowdsourced human labellers to further validate the accuracy of LLM labels. We worked with TicTag, a Singapore-based annotation company, to crowdsource for workers who could understand Singlish. These workers accessed the labelling task via TicTag's mobile application (see Appendix \ref{sec:tictag}), and were provided extensive instructions, including the safety risk taxonomy and examples. They could choose "I Don't Know" if they did not understand the text. 95 workers labelled 11,997 unique texts randomly drawn from our final dataset (see subsection \ref{sec:compiling-dataset}), with each text labelled by 3 different workers. The demographic profile of the workers were reflective of Singapore's population characteristics (see Appendix \ref{sec:tictag-workers}).

Of the 11,997 texts, we found that crowdsourced human labellers had low concurrence (i.e., inter-rater agreement). As seen in Table \ref{table:human-labeller-consensus}, human labellers only had full concurrence on binary labels 52.9\% of the time. Concurrence on binary labels was significantly lower than individual categories as we found that human labellers tended to label more texts \texttt{unsafe}, resulting in lower concurrence on generally safe texts. On less contentious categories like \texttt{self-harm}, \texttt{public harm}, \texttt{sexual} and \texttt{violence}, concurrence occurred more than 85\% of the time. In contrast, \texttt{toxic} and \texttt{hateful} categories had less than 75\% concurrence. Even with detailed instructions and strong quality control measures, the inherent subjectivity of labelling harmful content makes it challenging to achieve consensus among non-expert human labellers.


\begin{table}[h!]
\centering
 \begin{tabular}{m{2.1cm} m{2cm} m{2.3cm}} 
 \hline
 \textbf{Category} & \textbf{Human Consensus} & \textbf{Human-LLM Consensus}\\ [0.5ex] 
 \hline
 \texttt{hateful} & 70.6\% & 98.3\% (5,450)\\
 \texttt{harassment} & 82.0\% & 99.6\% (6,433)\\
 \texttt{public harm} & 87.9\% & 99.7\% (7,530)\\
 \texttt{self-harm} & 95.5\% & 100\% (6,817)\\
 \texttt{sexual} & 94.6\% & 99.8\% (4,234)\\
 \texttt{toxic} & 67.3\% & 97.8\% (7,475)\\
 \texttt{violent} & 94.3\% & 99.9\% (7,392)\\
 \hline
 \hline
\texttt{unsafe} & 52.9\% & 94.1\% (3,332)\\
 \hline
 \end{tabular}
\caption{Human consensus refers to full inter-rater agreement between human labellers. Human-LLM consensus refers to the consensus rate between human labellers and LLM labellers, with the number of texts in brackets. Note that only observations with full concurrence among all human labellers and LLM labellers for the respective categories were included in the latter, so the number varies depending on the category.}
\label{table:human-labeller-consensus}
\end{table}

For sentences with concurrence among all human labellers and all LLM labellers respectively, we found that the human labels generally have high concurrence with LLM labels (see Table \ref{table:human-labeller-consensus}), with the concurrence rate exceeding 90\% for all categories. This suggested that where human labels were consistent, LLMs were relatively accurate in providing labels aligned with human judgment. However, in contentious and ambiguous cases where human labels are inconsistent, evaluating the accuracy and concurrence of LLM labels vis-{\`a}-vis human labels is an area for future work.



\subsection{Classifier Results}
\label{sec:classifier-results}

\textbf{Model experimentation results} (see Table \ref{table:experimentation-results}): On the binary label, we found that the classifiers which used BGE Large performed significantly better than those which used HateBERT and SingBERT. Based on our ablation study with BERT-base, BERT-large and BGE-small models (see Appendix \ref{sec:experimentation-results-full}), which all performed poorly, we posit that the number of parameters and type of pre-training embeddings are critical in improving performance. As for the classifier model, the ridge classifier performed comparably to XGBoost and the neural network for all embedding models despite its relative simplicity. We also found that MLM finetuning on the embedding models had a negligible effect on performance (see Appendix \ref{sec:experimentation-results-full}).

For the multi-label classifiers, we similarly found that the classifiers which used BGE Large were the best performers by a large margin. Likewise, the ridge classifier performed best, indicating that a simple classification layer is sufficient for good performance, given a complex embedding model.


Overall, the best performing combination was the BGE model combined with the Ridge classifier. We used this combination for LionGuard, our moderation classifier.

\begin{table*}
\centering
\setlength{\cmidrulekern}{0.25em} 
\begin{tabular}{m{2.3cm} m{1.5cm} m{1cm} m{1cm} m{1cm} m{1cm} m{1cm} m{1cm} m{1cm} m{1cm}}
\toprule
  \multicolumn{2}{c}{\textbf{Moderation Classifier}} & \textbf{Binary} & \multicolumn{7}{c}{\textbf{Multi-Label}} \\
    \cmidrule(lr){1-2}
    \cmidrule(lr){3-3}
    \cmidrule(lr){4-10}
    \textbf{Embedding} & \textbf{Classifier} & \small\texttt{unsafe} & \small\texttt{hateful} & \small\texttt{harass-\newline ment} & \small\texttt{public harm} & \small\texttt{self-\newline harm} & \small\texttt{sexual} & \small\texttt{toxic} & \small\texttt{violent}\\
\hline

    & \textbf{Ridge} &
    \textbf{0.819} & \textbf{0.480} & \textbf{0.413} & \textbf{0.491} & \textbf{0.507} & \textbf{0.485} & \textbf{0.827} & \textbf{0.514}\\
    & XGBoost & 0.816
    & 0.455 & 0.386 & 0.460 & 0.472 & 0.472 & 0.807 & 0.489 \\
    \multirow{-3}{3cm}{\textbf{BGE Large}} & NN & 
    0.792 & 0.375 & 0.254 & 0.319 & 0.286 & 0.388 & 0.802 & 0.299 \\
    \hline

    \multirow{3}{3cm}{HateBERT} & Ridge & 0.083 & 0.065 & 0.063 & 0.068 & 0.079 & 0.064 & 0.076 & 0.066\\
    & XGBoost & 0.082 & 0.064 & 0.064 & 0.067 & 0.078 & 0.064 & 0.073 & 0.064 \\
    & NN & 0.082 & 0.064 & 0.059 & 0.063 & 0.073 & 0.063 & 0.073 & 0.059 \\
    \hline
    
    \multirow{3}{3cm}{SingBERT} & Ridge & 0.194 & 0.121 & 0.119 & 0.131 & 0.139 & 0.114 & 0.186 & 0.125\\
    & XGBoost & 0.172 & 0.112 & 0.099 & 0.115 & 0.119 & 0.103 & 0.167 & 0.111\\
    & NN & 0.155 & 0.090 & 0.061 & 0.067 & 0.074 & 0.063 & 0.123 & 0.063 \\
    \hline

    \multicolumn{2}{c}{Moderation API} & 0.675 & 0.228 & 0.081 & - & 0.488 & 0.230 & - & 0.137\\
    \hline

    \multicolumn{2}{c}{Perspective API} & 0.588 & 0.212 & 0.126 & - & - & - & 0.342 & 0.073\\
    \hline

    \multicolumn{2}{c}{LlamaGuard} & 0.459 & 0.190 & - & 0.031  & 0.370 & 0.230 & - & 0.005\\
    
\bottomrule
\end{tabular}
\caption{Comparison of PR-AUC between different combinations of embedding and classifier models for the binary label (safe or unsafe) and the seven safety risk categories against Moderation API, Perspective API and LlamaGuard. The top score for each category is formatted in bold for clarity, and the combination used for LionGuard is in bold. The full table (including results from our finetuned embedding models) is available in Appendix \ref{table:experimentation-results-full}}.
\label{table:experimentation-results}
\end{table*}

\textbf{Benchmarking results} (see Table \ref{table:experimentation-results}): We found that LionGuard significantly outperformed Moderation API, Perspective API and LlamaGuard.

On the binary label experiments, LionGuard's PR-AUC score of 0.819 is higher than OpenAI's 0.675, Perspective's 0.588 and LlamaGuard's 0.459.\footnote{Additionally, LionGuard scored lower than Moderation API on precision (0.63 vs 0.74) but significantly higher on recall (0.81 vs 0.56) when using 0.5 as the prediction threshold.} Likewise, for multi-label classification, LionGuard outperformed on all categories. The difference in performance is especially clear for the \texttt{harassment}, \texttt{sexual}, \texttt{toxic} and \texttt{violent} categories, with the performance more than doubled. 

\section{Discussion}
\textbf{Importance of localization}: Our work suggests a clear need for contextualized moderation classifiers to detect localized slang and dysphemisms that are not offensive elsewhere. In our error analysis of a few examples where Moderation API, Perspective API and LlamaGuard failed to provide accurate labels (see Table \ref{table:example-predictions} in Appendix \ref{sec:appendix-example-predictions}), LionGuard was able to understand Singapore-specific slang and references like "\textit{ceca}", "\textit{kkj}" and "\textit{AMDK}" and provide the correct label. In contrast, Moderation API, Perspective API and LlamaGuard seemed to perform better in examples where only offensive English words or references (e.g. "\textit{leeches}", "\textit{wank}", "\textit{scum}") were present. Hence, while Moderation API, Perspective API and LlamaGuard are well-adapted to Western-centric toxicity, LionGuard is able to perform better on Singlish texts. 

However, LionGuard may not generalize well to other domains and languages, as it was trained specifically to detect harmful content in the Singapore context. Nonetheless, our approach can be adapted to any low-resource languages which require localization. Future work can use LionGuard to generate adversarial Singlish-centric data \citep{perez} to augment moderation training data and refine generalist moderation classifiers \citep{openaimoderation} for better performance even on low-resource languages.

\textbf{Benefits of automated LLM labelling}: While crowdsourced labelling works well with simple tasks with an objective truth, we found that it may have limited mileage for subjective tasks like assessing toxicity or harassment. Each person has a different understanding of what is unsafe, and aligning on these definitions is challenging. On the other hand, automated LLM labelling, with the right prompt, can achieve higher labelling accuracy and consistency. More importantly, this approach can be adapted to other low-resource languages, and easily updated as the language evolves. While our work adopted the consensus approach for automated labelling, future work can explore other methods for synthesizing varying LLM output labels, including self-reflection.

\section{Conclusion}

We highlighted the importance of low-resource language localization for moderation by showing that our finetuned classifier, LionGuard, outperformed existing widely-used moderation APIs. We evaluated the best prompt and LLMs for automatic labelling, and presented a practical and scalable approach to automatically generating labels for low-resource language moderation data. We hope our work encourages more to build moderation tools that excel in both general and localized contexts.

\pagebreak

\section{Limitations}
\textbf{Dataset.} As our dataset is a static, albeit up-to-date, snapshot of the online discourse in Singapore, our model may become less effective as the linguistic features of Singlish inevitably change over time. However, our approach simply requires a change in the prompt used for labelling in order to generate accurate labels for a new dataset. Moreover, active learning can be used to continually learn from production data \citep{openaimoderation} and ensure that the moderation classifier performs well over time. Future work can incorporate our methods (e.g., automated labelling) into an end-to-end pipeline to ensure a robust detection model for real-world applications. 

\textbf{Experiments.} As the focus of our work was to highlight the importance of localized content moderation, we did not perform extensive experiments on varying model hyperparameters. However, our results found that the ridge classifier, which only depended on one hyperparameter, could outperform generalist classifiers. For future work, we hope to experiment with finetuning encoder-decoder transformer language models with a classification output layer, which may perform even better than our relatively simple classifier models.

\textbf{LLM Labelling Accuracy.} While we have attempted to validate our LLM labels both with our expert-labelled dataset and crowd-sourced labels, we cannot completely guarantee the accuracy of our LLM labels. However, our work aims to demonstrate the potential of LLM labelling with prompt engineering, as an alternative to manual human-labelling. Future work can explore more advanced LLM labelling techniques to increase accuracy. 

\section{Ethical Considerations}

\textbf{Labeller Wellbeing.} Workers were informed about the nature of the task before commencing their work. They completed their work in batches, on their own schedules, and could decide to withdraw at any point in time. Trigger warnings were placed in the task description and mental health resources were made available by TicTag to the workers. Workers were compensated at a rate of SG\$0.20 per text annotated. TicTag shared that the workers annotated approximately 80 texts per half an hour, which adds up to SG\$32 per hour, well above the living wage in Singapore. No identifiable information was provided to us about our workers. 

\textbf{Data Privacy and Terms of Use.} Reddit data was collected via the Pushshift API \citep{pushshift}. We collected Hardwarezone data that was publicly available, in a manner that is permissible pursuant to the Singapore Copyright Act 2021, which allows for the use of copyrighted works for computational data analysis (i.e., machine learning). 

\textbf{Model Terms of Use.} We used LLMs commercially licensed by OpenAI, Anthropic and Google and abided by their Terms of Use. We also accessed Llama 2 via Hugging Face, licensed by Meta. We accepted and abided by Meta's license terms and acceptable use policy. We accessed BGE, SingBERT and HateBERT via Hugging Face Hub and abided by their Terms of Use. Our moderation classifier, LionGuard, will be made available on Hugging Face for research and public interest purposes only.

\textbf{Environmental Impact.} We only trained lightweight models in our main experiments, such as a ridge classifier, XGBoost and a simple neural network. The most significant training required was unsupervised MLM fine-tuning of the embedding models, which took approximately three days on two NVIDIA Tesla V100s. Compared to the environmental costs of pre-training LLMs, the environmental impact of our work is relatively small. 


\bibliography{custom}

\pagebreak

\appendix

\section{List of Controversial Topics and Words}
\label{sec:controversial-topics}

"ceca", "ghey", "tiong", "abnn", "amdl", "amdk", "pinoy", "jiuhu", "prc", "indian", "filipino", "foreign", "angmo", "spg", "atb", "chennai", "****", "bbm", "ft", "fw", "transformer", "chink", "bangla", "yalam", "curry", "piak", "syt", "fap", "pcc", "nnp", "pika", " kkj", "abalone", "asgm", "btss", "hmv", "humsup", "milf", "nekkid", "nsfw", "ocb", "okt", "pcc", "perbird", "tps", "vpl", "parang", "slash", "punch", "kick", "shoot", "buibui", "bbfa", "cheesepie", "gcp", "diu lei", "ccb", "siao", "cheese pie", "knn", "ccb", "pcb", "smlj", "tiu", "rcp", "asw", "bus3rd", "digger", "gcp", "vape", "weed", "drug", "launder", "wash money", "377a", "raeesah khan", "oxley", "halimah", "brownface", "chinese privilege", "presidential election", "ashlee", "wuhan", "mahathir", "pink dot", "egg freezing", "kong hee", "schooling", "amos yee", "kurt tay"

\section{Examples of Singlish texts in our dataset}
\label{sec:singlish-examples}

\begin{itemize}[nolistsep]
    \item \textit{"Hahahah now i dont play nice liao and take longer liao but my personality cannot sia hahahaha dont like things to be left hanging there. Now not my daiji i will throw to others hais but company cut cost so i got extra job too"} (Comment from Reddit, posted on Oct 2020)
    
    \item \textit{"walao, never share with us, that's why you kena caught lah"} (Comment from Reddit, posted on May 2021)
    
    \item \textit{"wah lao i jio then u say cut down. knn jin hurt my heart. price is ok i will share with my friends"} (Comment from HardwareZone, posted on Sep 2022)
    
    \item \textit{"siao bo, write such a long WOT to "prove" SAP ppl are mostly jiak kantangs. WTF"} (Comment from HardwareZone, posted on Feb 2023)
\end{itemize}

\section{Toxicity Categories}
\label{sec:app-toxicity-categories}
\begin{enumerate}[nolistsep]
    \item \textbf{Hateful}: Content that expresses, incites, or promotes hate based on race, gender, ethnicity, religion, nationality, sexual orientation, disability status, or caste. Hateful content aimed at non-protected groups (e.g., chess players) is harrassment. Hateful content that also includes violence or serious harm towards the targeted group based on race, gender, ethnicity, religion, nationality, sexual orientation, disability status, or caste. 
    \item \textbf{Harassment}: Content that expresses, incites, or promotes harassing language towards any target/individual. Without mention of violence or harm, content that results in prolonged mental or emotional suffering for the target, enduring for more than an hour. Any harassment content that includes violence or serious harm towards any target/individual. 
    \item \textbf{Encouraging public harm}: Content that promotes, facilitates, or encourages harmful public acts, vice or organized crime.
    \item \textbf{Encouraging self-harm}: Content that promotes, encourages, or depicts acts of self-harm, such as suicide, cutting, and eating disorders. Content that encourages performing acts of self-harm, such as suicide, cutting, and eating disorders, or that gives instructions or advice on how to commit such acts.
    \item \textbf{Sexual}:  Content meant to arouse sexual excitement, such as the description of sexual activity, or that promotes sexual services (excluding sex education and wellness). Sexual content that includes an individual who is under 18 years old.
    \item \textbf{Toxic}: Content that is rude, disrespectful, or profane, including the use of slurs. A very hateful, aggressive, disrespectful comment or otherwise very likely to make a user leave a discussion or give up on sharing their perspective.
    \item \textbf{Violent}: Content that depicts death, violence, or physical injury. 
\end{enumerate}

\section{Labelling prompt}
\label{sec:app-prompt}

We share the full prompt used for automated labelling below, highlighting in different colors which sections correspond to context prompting, few-shot prompting, and Chain-of-Thought prompting respectively. 
\newline \newline
\textcolor{brown}{\textbf{Context Prompting}}
\newline
\textcolor{brown}{\texttt{
Context: You are helping to identify if a given Singlish comment is toxic by understanding its meaning and evaluating it on seven criteria.\\
\\
Some examples of Singaporean slang and their definitions are provided below.\\
ghey: male homosexual, usually in a derogatory way\\
tiong: people from China, may be in a derogatory way\\
ceca: people from India, usually in a derogatory way\\
bbfa: people who are overweight and resigned to being alone\\
yalam: people who are ethnically Malay\\
fap: masturbation\\
piak: sexual intercourse\\
syt: sexy young thing\\
ZG or zao geng: accidentally showing underwear in public, usually for women\\
nnp: nipples\\
amdk: sarcastic reference to Caucasians being superior\\
siao: crazy\\
knn: vulgar word\\
kns: vulgar word\\
sot: Singlish for "shot", meaning no longer functional\\}}

{\texttt{Task: Iterate through the list of seven criteria below and determine whether the Singlish comment below, which includes Singaporean terms, fulfil any of the seven criteria below. You should be a more liberal labeller - if something has a 50\% chance of being toxic, you should label it as such. You also need to take into account the Singlish terms, vulgarities, and unique racial, religious, and cosmopolitan context.\\
a. hateful: Content that expresses, incites, or promotes hate based on race, gender, ethnicity, religion, nationality, sexual orientation, disability status, or caste. Hateful content aimed at non-protected groups (e.g., chess players) is harrassment. Hateful content that also includes violence or serious harm towards the targeted group based on race, gender, ethnicity, religion, nationality, sexual orientation, disability status, or caste. \\
b. harassment: Content that expresses, incites, or promotes harassing language towards any target/individual. Harassment content that also includes violence or serious harm towards any target/individual. \\
c. encouraging public harm: Content that promotes, facilitates, or encourages harmful public acts, vice or organised crime.\\
d. encouraging self-harm: Content that promotes, encourages, or depicts acts of self-harm, such as suicide, cutting, and eating disorders. Content that encourages performing acts of self-harm, such as suicide, cutting, and eating disorders, or that gives instructions or advice on how to commit such acts.\\
e. sexual: Content meant to arouse sexual excitement, such as the description of sexual activity, or that promotes sexual services (excluding sex education and wellness). Sexual content that includes an individual who is under 18 years old.\\
f. toxic: Content that is rude, disrespectful, or profane, including the use of slurs. A very hateful, aggressive, disrespectful comment or otherwise very likely to make a user leave a discussion or give up on sharing their perspective.
g. violent: Content that depicts death, violence, or physical injury.\\}

\textcolor{blue}{\textbf{Chain-of-Thought Prompting}}
\newline
\textcolor{blue}{\texttt{For each criterion, do the following:\\
1. Consider the Singlish comment below and determine whether the comment fulfills the criterion.\\
2. Provide a "Yes" or "No" label.\\
3. Provide a reason for the label.\\
4. Output the answers strictly in the following JSON format, replacing <criterion> with the criteria listed below:\\
\{\\
\indent"<criterion>": \{\\
\indent\indent  "label": "Yes" / "No",\\
\indent\indent  "reason": "<reason>"\\
\indent    \}\\
\}\\}}

\textcolor{purple}{\textbf{Few-shot Prompting}}
\newline
\textcolor{purple}{\texttt{Here are some examples of Singlish comments and which criteria they fulfill.\\
\\
Singlish comment: "I thinking of making a school girl ZG/upskirt video, something like the schoolgirl above, then open legs show panties, or even pulling up her skirt let ppl beo. but donno will kena complain anot. Youtube alot of cheesepie ppl like to report for very minor things..."\\
Criteria fulfilled: encouraging public harm, sexual, toxic\\
\\
Singlish comment: "yes. I want the judge to sentence the max jail term and caning. plus a bonus of they have to rape each other monthly. The loser will have to be rape again by another inmate."\\
Criteria flagged: harassment, toxic, encouraging public harm, sexual\\
\\
Singlish comment: "Hi ib, u 1 day don't kanna fug u don't feel good si bo? Knnbccb"\\
Criteria flagged: toxic\\
\\
Singlish comment: "wah piang eh there ish something insai every man dream on piaking girls in school uniform mens"\\
Criteria flagged: sexual\\
\\
Singlish comment: ""nowadays pple dont have much integrity 1 la wont suicide 1. they will treat it lightly and say so be it"\\
Criteria flagged: encouraging self harm\\
\\
Singlish comment: "Pray tell me why ceca cannot break through in other Asian economies like Korea, Japan, Taiwan, China if they are so good and compete fair and square? They only thrive in Singapore simply because ceca and our gahmen close 1 eye to their unfair hiring. By way I not pro singkee, since I migrate. Juz 2 cents"\\
Criteria flagged: hateful\\
\\
Singlish comment: "The beach road tiong guy slashed his wife during broad daylight with full intention of killing her. Even severing her hand off and blinded her in one eye. It is over relationship issue as well like this case."\\
Criteria flagged: violent\\}}

\texttt{Remember to return the output in a JSON format. The Singlish comment for your evaluation is: <insert statement here>\\}}

\section{Error Analysis of Llama 2 Predictions}
\label{sec:llama2-error-analysis}
See Table \ref{table:llama2-error-analysis} for examples in which Llama 2 inaccurately classified safe texts as unsafe.

\begin{table}[h!]
\centering
 \begin{tabular}{m{3.5cm} m{3.5cm}} 
 \hline
 \textbf{text} & \textbf{Output} \\ [0.5ex] 
 \hline
where got po chai pills in the original article... ya loh, scare me sia. i always carry them for my travels. & \{..."sexual": \{"label": "Yes", "reason": "The comment mentions 'po chai pills' which could be interpreted as a reference to sexual enhancement drugs, and the commenter mentions carrying them for travels, which could be seen as a suggestion of sexual activity."\}, "toxic": \{"label": "Yes", "reason": "The comment contains a mild profanity 'ya loh' and the phrase 'scare me sia' which could be seen as a form of intimidation or threat."\}...\}\\
\hline
Aiyaa shd just jitao go on dating apps if he really wants to find his dream gal rahz & \{..."sexual": \{"label": "Yes", "reason": "The comment contains sexual content meant to arouse sexual excitement, such as the description of sexual activity."\}, "toxic": \{"label": "Yes", "reason": "The comment contains rude, disrespectful, or profane language, including the use of slurs."\}...\} \\
\hline
Moi c moi company the data science team oso use tableau click here click there, earn 1.5 times of moi celery & \{..."toxic": \{"label": "Yes", "reason": "The comment contains derogatory language and disrespectful tone."\}...\}\\
\hline
 \end{tabular}
\caption{Examples of safe texts labelled unsafe by Llama 2 using full prompt.}
\label{table:llama2-error-analysis}
\end{table}

\section{Final set of hyperparameters}
\label{sec:hyperparameter-results}

\subsection{Ridge Classifier}
  
\textit{Hyperparameters}:
\begin{itemize}[nolistsep]
  \item $\alpha$: 1.0
\end{itemize}

\textit{Hyperparameter search}: No hyperparameter search was conducted on the ridge classifier as it performed well out of the box. 

\subsection{XGBoost}

\textit{Hyperparameters}:
\begin{itemize}[nolistsep]
  \item \texttt{max depth}: 6
  \item \texttt{learning rate}: 0.2
  \item \texttt{scale pos weight}: 5
  \item \texttt{n estimators}: 5
\end{itemize}

\textit{Hyperparameter search}: A halving grid search and a standard grid search was performed on all the parameters listed above, with 5-fold cross validation on the training set and evaluation on the validation set.

\subsection{Neural Network}

\textit{Hyperparameters}:
\begin{itemize}[nolistsep]
  \item \texttt{epochs}: 30
  \item \texttt{batch size}: 8
  \item \texttt{learning rate}: 0.001
\end{itemize}

\textit{Hyperparameter search}: A halving grid search and a standard grid search was performed on all the parameters listed above, with evaluation on the validation set.

\section{TicTag Labelling}
\label{sec:tictag}

\subsection{Crowd-sourced Workers Profiles}
\label{sec:tictag-workers}
Of the 95 crowd-sourced workers, 89\% were Chinese, 5\% were Malay, 3\% were Indian and 1\% were Other. 47\% of workers were aged 18-24, 31\% were aged 24-34, 15\% were aged 35-44 and the remaining 4\% were aged 45-54. 53\% of workers were female, while the remaining 44\% were male. Workers were all residents of Singapore. 

\subsection{Annotation Interface}

TicTag designed the following mobile application interface to obtain crowd-sourced annotations. Instructions were provided in English, but some button options were provided in chosen native languages. We show screenshots of the interface in Malay. 
\begin{figure}[H]
\centering
    \includegraphics[width=0.7\linewidth]{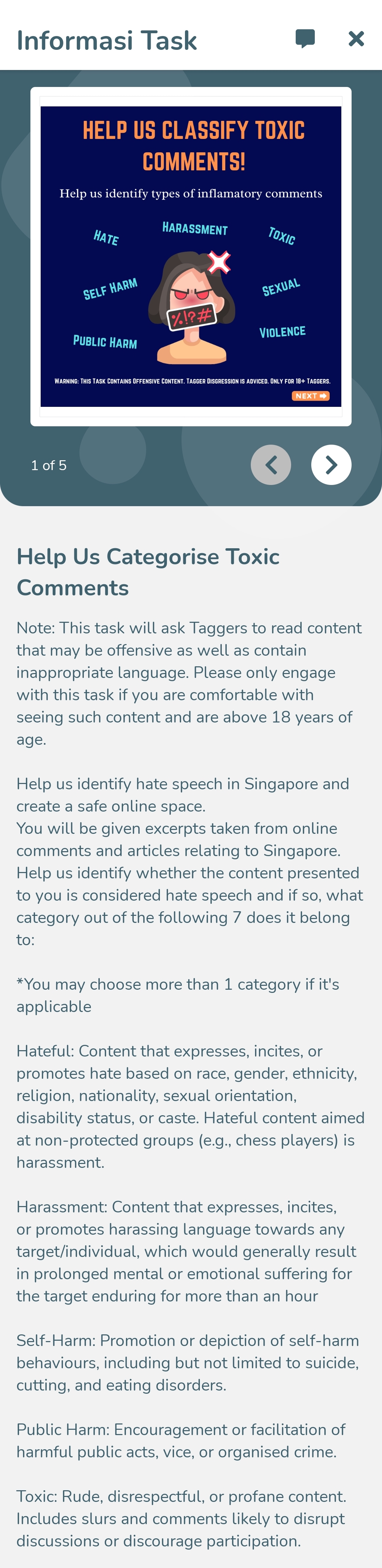}
\caption{Instructions Page. Page 1 of top section shows generic task title descriptions. Bottom section is a scrollable section that shows detailed task description and trigger warning.}
\label{fig:tictag-p1}
\end{figure}

\begin{figure}[H]
\centering
    \includegraphics[width=0.7\linewidth]{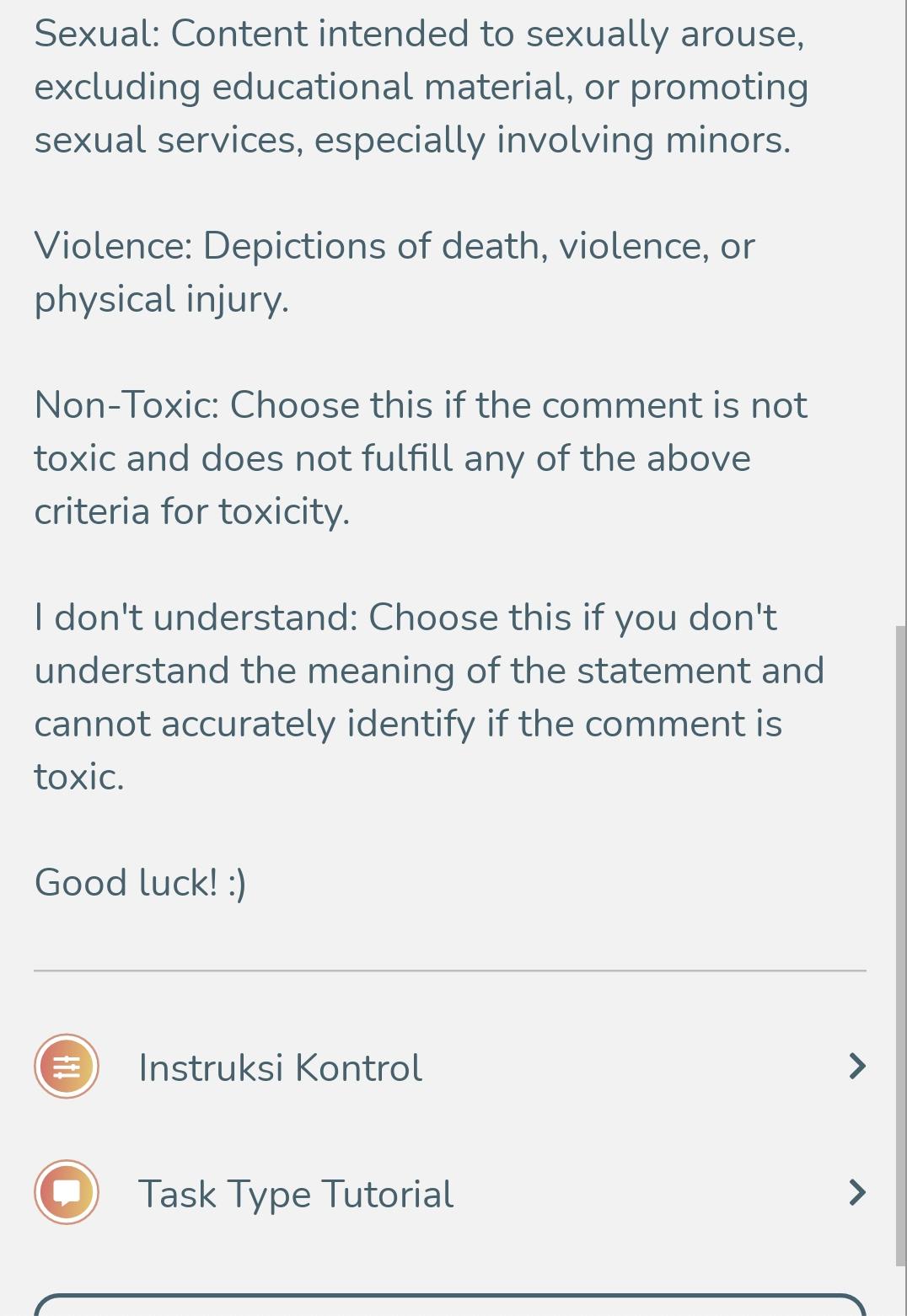}
\caption{Instructions Page. Continued bottom section from Fig \ref{fig:tictag-p1} contains descriptions of remaining safety risk category and the "I Don't Know" option for the workers.}
\end{figure}

\begin{figure}[H]
\centering
    \includegraphics[width=0.7\linewidth]{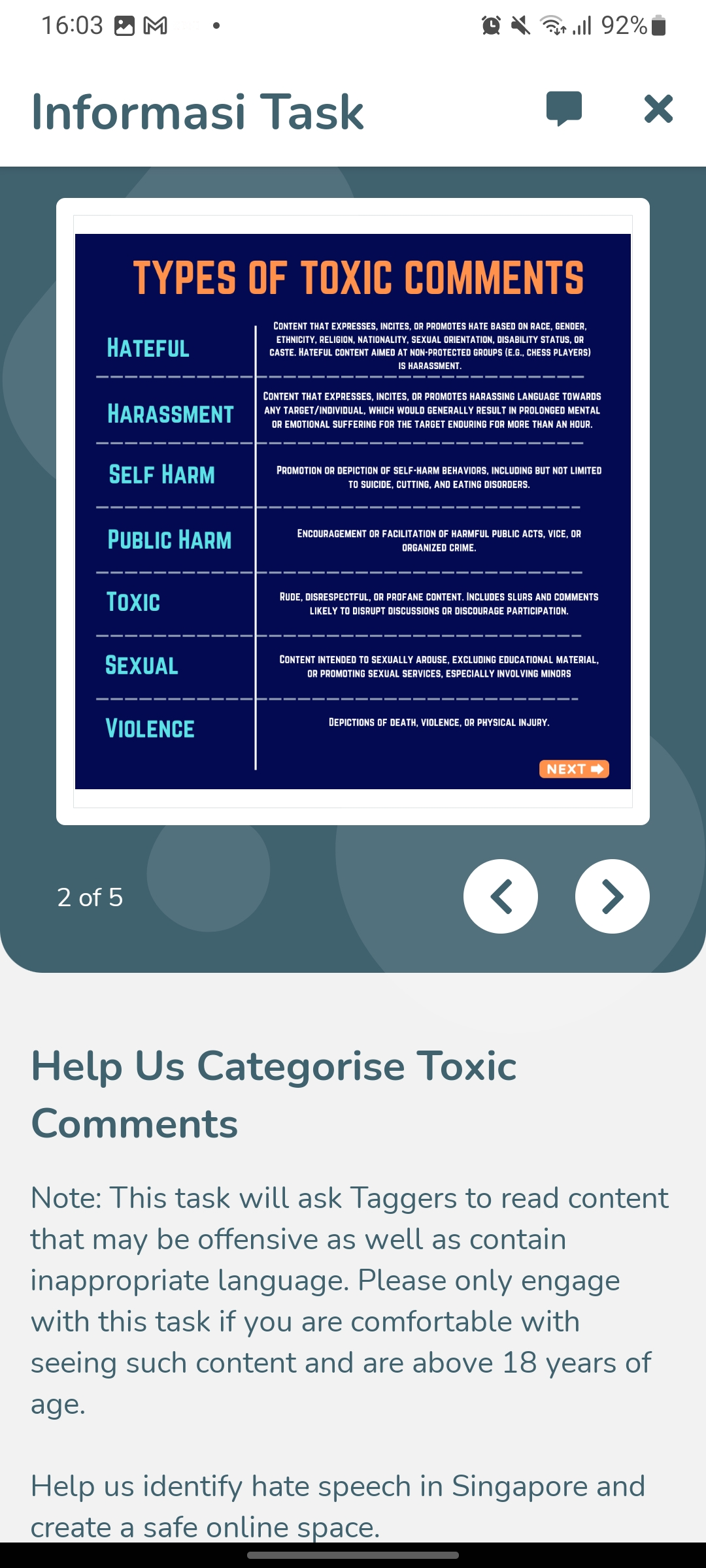}
\caption{Instructions Page. Page 2 of the top section.}
\end{figure}

\begin{figure}[H]
\centering
    \includegraphics[width=0.7\linewidth]{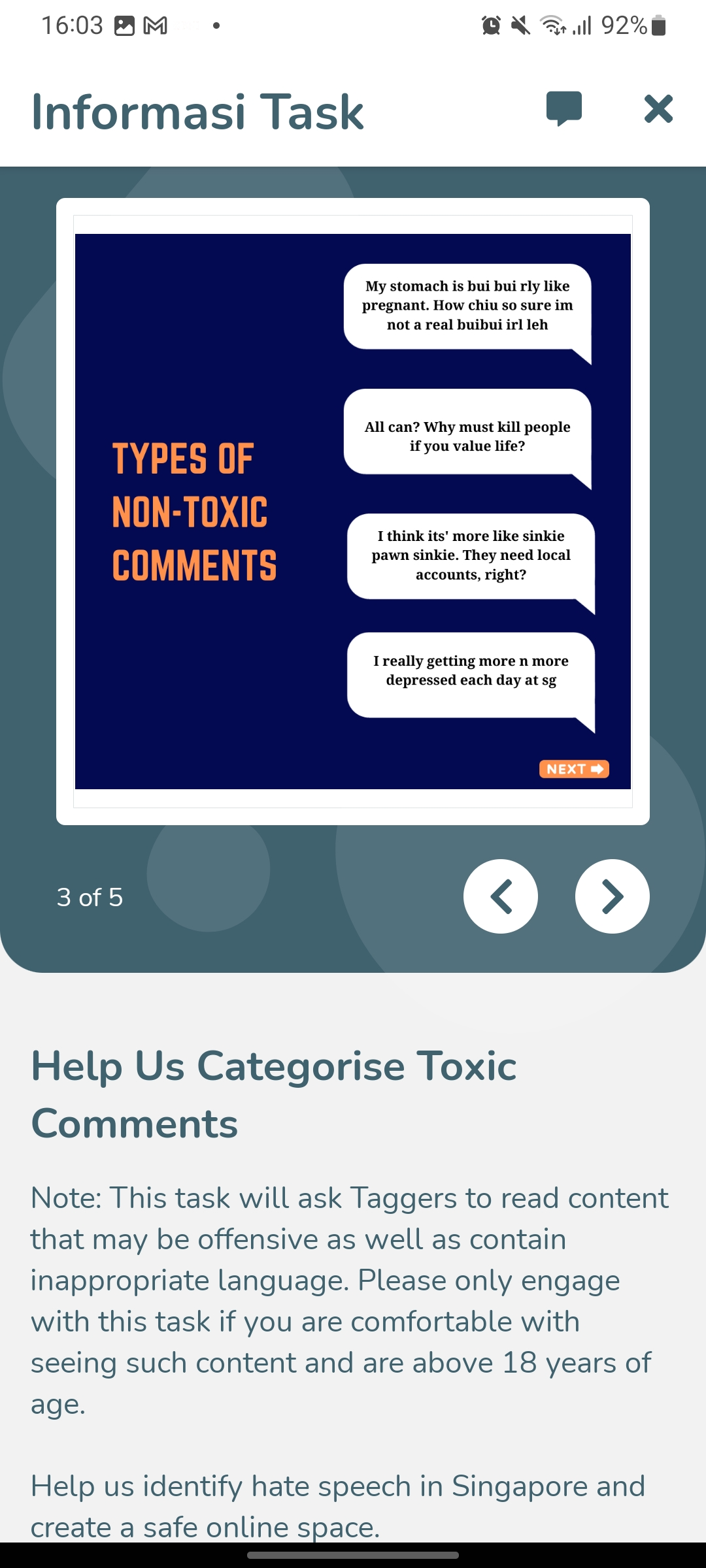}
    \includegraphics[width=0.7\linewidth]{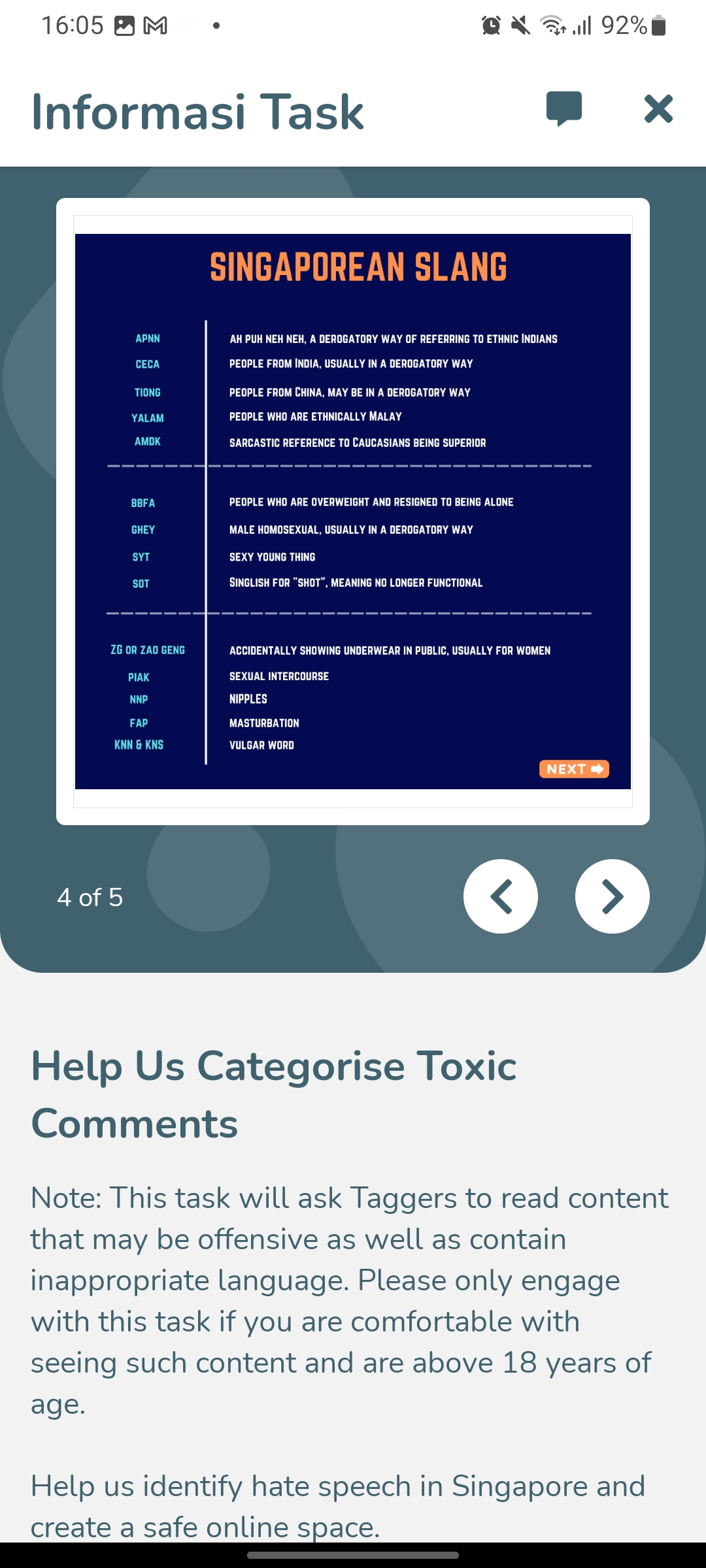}
\caption{Instructions Page. Pages 3-4 of the top section.}
\end{figure}

\begin{figure}[H]
\centering
    \includegraphics[width=0.65\linewidth]{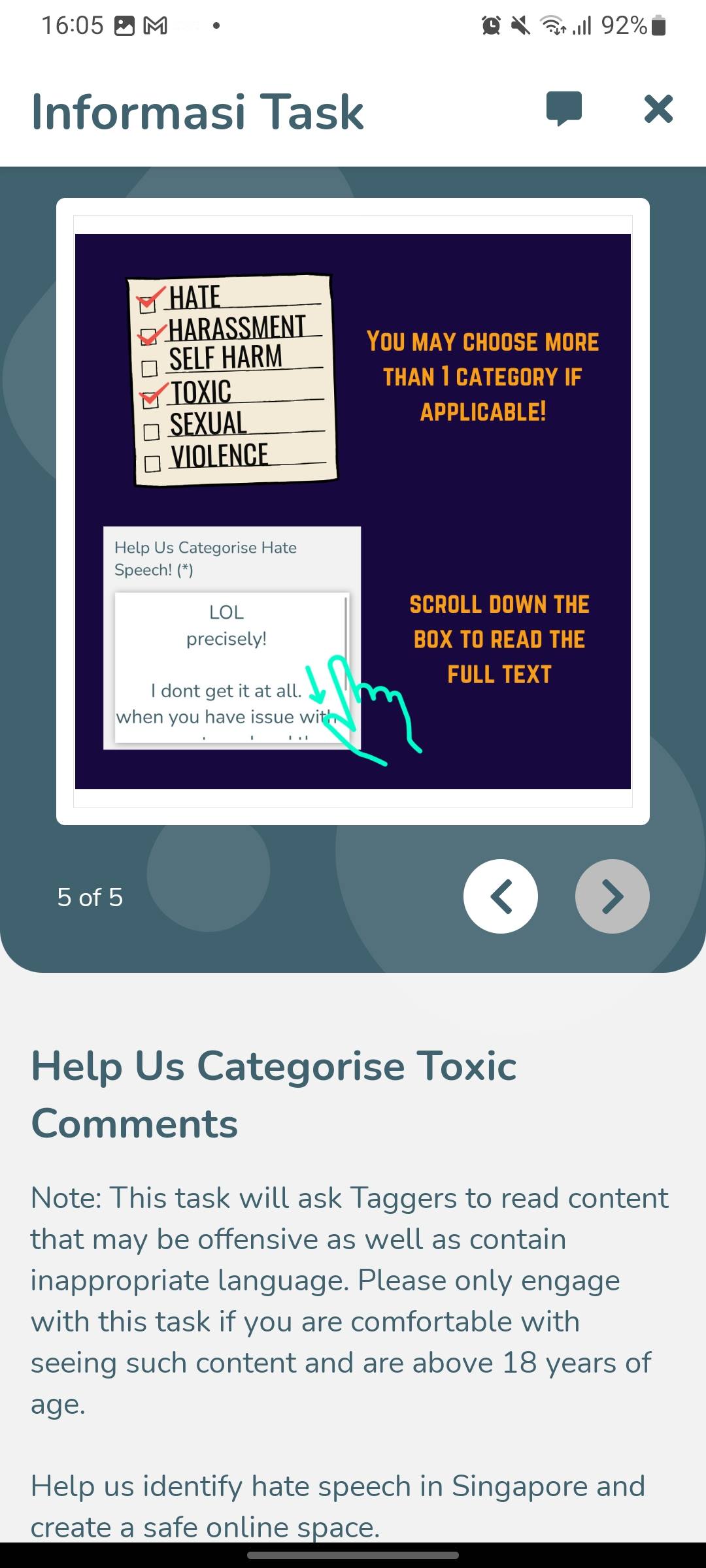}
\caption{Instructions Page. Page 5 of the top section.}
\end{figure}

\begin{figure}[H]
\centering
    \includegraphics[width=0.65\linewidth]{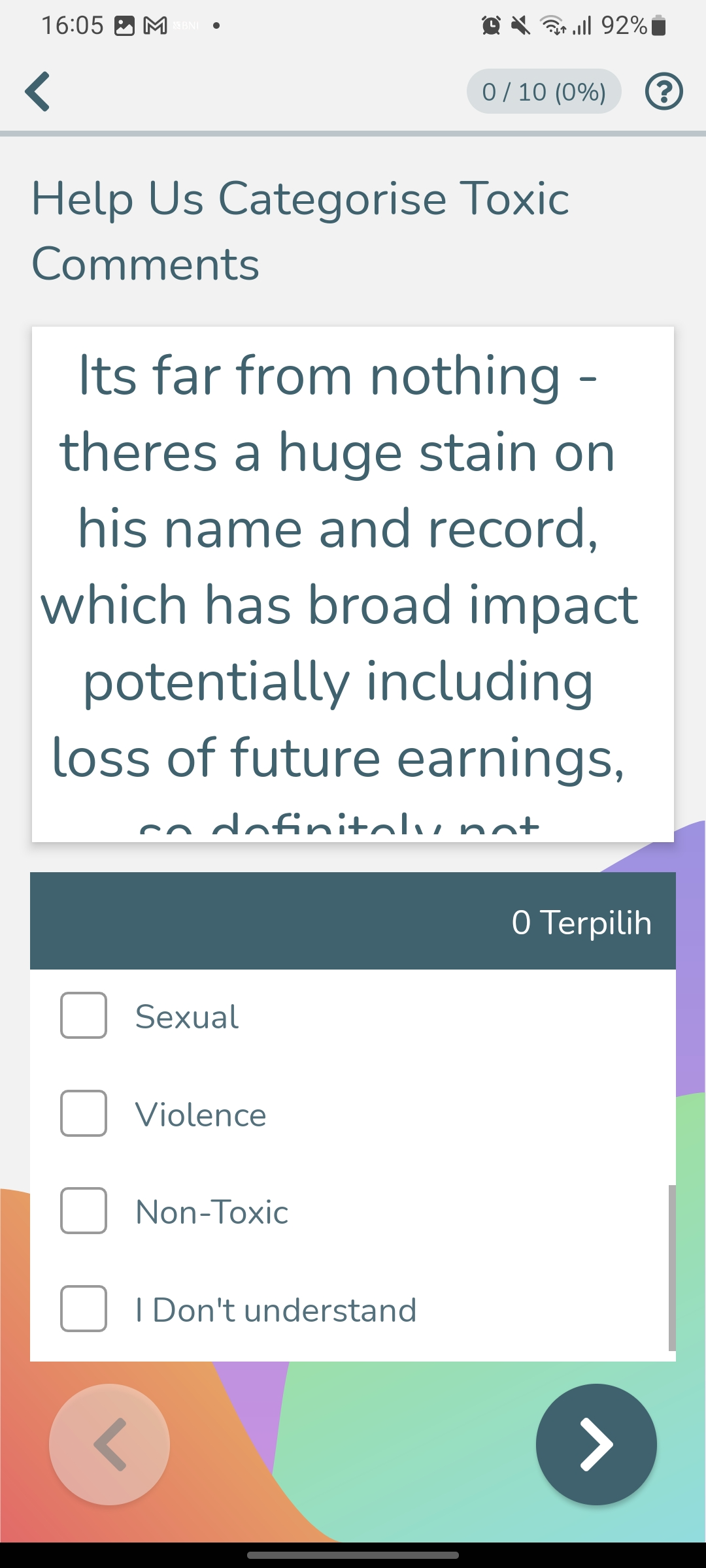}
\caption{Annotation Page with labelling actions.}
\end{figure}

\section{Full experimentation results}
\label{sec:experimentation-results-full}

See Table \ref{table:experimentation-results-full} in the next page for the full comparison of all experimentation and benchmarking results.

\begin{table*}[h!]
\centering
\setlength{\cmidrulekern}{0.25em} 
\begin{tabular}{m{2.3cm} m{1.5cm} m{1cm} m{1cm} m{1cm} m{1cm} m{1cm} m{1cm} m{1cm} m{1cm}}
\toprule
  \multicolumn{2}{c}{\textbf{Moderation Classifier}} & \textbf{Binary} & \multicolumn{7}{c}{\textbf{Multi-Label}} \\
    \cmidrule(lr){1-2}
    \cmidrule(lr){3-3}
    \cmidrule(lr){4-10}
    \textbf{Embedding (\# parameters)} & \textbf{Classifier} & \small\texttt{unsafe} & \small\texttt{hateful} & \small\texttt{harass-\newline ment} & \small\texttt{public harm} & \small\texttt{self-\newline harm} & \small\texttt{sexual} & \small\texttt{toxic} & \small\texttt{violent}\\
\midrule

    & \textbf{Ridge} & \textbf{0.819} & \textbf{0.480} & \textbf{0.413} & \textbf{0.491} & \textbf{0.507} & \textbf{0.485} & \textbf{0.827} & \textbf{0.514}\\
    & XGBoost & 0.816 & 0.455 & 0.386 & 0.460 & 0.472 & 0.472 & 0.807 & 0.489 \\
    \multirow{-3}{2.3cm}{\textbf{BGE Large} (326m)} & NN & 0.792 & 0.375 & 0.254 & 0.319 & 0.286 & 0.388 & 0.802 & 0.299 \\
    \hline

    \multirow{3}{2.3cm}{HateBERT (110m)} & Ridge & 0.083 & 0.065 & 0.063 & 0.068 & 0.079 & 0.064 & 0.076 & 0.066\\
    & XGBoost & 0.082 & 0.064 & 0.064 & 0.067 & 0.078 & 0.064 & 0.073 & 0.064 \\
    & NN & 0.082 & 0.064 & 0.059 & 0.063 & 0.073 & 0.063 & 0.073 & 0.059 \\
    \hline
    
    \multirow{3}{2.3cm}{SingBERT (110m)} & Ridge & 0.194 & 0.121 & 0.119 & 0.131 & 0.139 & 0.114 & 0.186 & 0.125\\
    & XGBoost & 0.172 & 0.112 & 0.099 & 0.115 & 0.119 & 0.103 & 0.167 & 0.111\\
    & NN & 0.155 & 0.090 & 0.061 & 0.067 & 0.074 & 0.063 & 0.123 & 0.063 \\
    \hline
    
    \multirow{3}{3cm}{BGE Large finetuned (326m)} & Ridge & 0.794 & 0.466 & 0.402 & 0.464 & 0.474 & 0.455 & 0.794 & 0.498\\
    & XGBoost & 0.789 & 0.461 & 0.386 & 0.444 & 0.448 & 0.438 & 0.777 & 0.452 \\
    & NN & 0.771 & 0.357 & 0.277 & 0.304 & 0.275 & 0.343 & 0.781 & 0.348 \\
    \hline

    \multirow{3}{3cm}{HateBERT finetuned (110m)} & Ridge & 0.187 & 0.120 & 0.122 & 0.127 & 0.137 & 0.117 & 0.178 & 0.125\\
    & XGBoost & 0.172 & 0.112 & 0.099 & 0.116 & 0.121 & 0.104 & 0.167 & 0.112 \\
    & NN & 0.134 & 0.088 & 0.061 & 0.066 & 0.074 & 0.075 & 0.133 & 0.062 \\
    \hline

    \multirow{3}{3cm}{SingBERT finetuned (110m)} & Ridge & 0.191 & 0.122 & 0.117 & 0.132 & 0.137 & 0.115 & 0.186 & 0.125\\
    & XGBoost & 0.172 & 0.112 & 0.099 & 0.116 & 0.120 & 0.103 & 0.167 & 0.111 \\
    & NN & 0.145 & 0.060 & 0.065 & 0.067 & 0.074 & 0.084 & 0.143 & 0.063 \\
    \hline

    & Ridge & 0.183 & 0.120 & 0.114 & 0.127 & 0.135 & 0.113 & 0.179 & 0.125 \\ 
    & XGBoost & 0.174 & 0.112 & 0.098 & 0.116 & 0.120 & 0.103 & 0.168 & 0.112 \\
    \multirow{-3}{2.3cm}{BERT Large (340m)} & NN & 0.152 & 0.087 & 0.062 & 0.067 & 0.074 & 0.087 & 0.118 & 0.062 \\ \hline

    & Ridge & 0.178 & 0.057 & 0.004 & 0.007 & 0.001 & 0.022 & 0.172 & 0.001 \\
    & XGBoost & 0.176 & 0.112 & 0.098 & 0.116 & 0.121 & 0.103 & 0.167 & 0.112 \\
    \multirow{-3}{2.3cm}{BERT Base (110m)} & NN & 0.139 & 0.060 & 0.062 & 0.066 & 0.073 & 0.074 & 0.127 & 0.063 \\ 
    \hline

    & Ridge & 0.171 & 0.116 & 0.113 & 0.126 & 0.132 & 0.108 & 0.166 & 0.120 \\ 
    & XGBoost & 0.175 & 0.113 & 0.099 & 0.116 & 0.121 & 0.104 & 0.167 & 0.112 \\
    \multirow{-3}{2.3cm}{BGE Small (24m)} & NN & 0.138 & 0.093 & 0.062 & 0.067 & 0.074 & 0.067 & 0.131 & 0.063 \\ \hline

    \multicolumn{2}{c}{Moderation API} & 0.675 & 0.228 & 0.081 & - & 0.488 & 0.230 & - & 0.137\\
    \hline

    \multicolumn{2}{c}{Perspective API} & 0.588 & 0.212 & 0.126 & - & - & - & 0.342 & 0.073\\
    \hline

    \multicolumn{2}{c}{LlamaGuard} & 0.459 & 0.190 & - & 0.031  & 0.370 & 0.230 & - & 0.005\\
    
\bottomrule
\end{tabular}
\caption{Comparison of PR-AUC between different combinations of embedding (including finetuned ones) and classifier models for the binary label (safe or unsafe) and the seven safety risk categories against Moderation API, Perspective API and LlamaGuard. The top score for each category is formatted in bold for clarity, and the combination used for LionGuard is in bold.}
\label{table:experimentation-results-full}
\end{table*}

\begin{table*}[h!]
\centering
\setlength{\cmidrulekern}{0.25em} 
\begin{tabular}{m{2.3cm} m{1.5cm} m{1cm} m{1cm} m{1cm} m{1cm} m{1cm} m{1cm} m{1cm} m{1cm}}
\toprule
  \multicolumn{2}{c}{\textbf{Moderation Classifier}} & \textbf{Binary} & \multicolumn{7}{c}{\textbf{Multi-Label}} \\
    \cmidrule(lr){1-2}
    \cmidrule(lr){3-3}
    \cmidrule(lr){4-10}
    \textbf{Embedding} & \textbf{Classifier} & \small\texttt{unsafe} & \small\texttt{hateful} & \small\texttt{harass-\newline ment} & \small\texttt{public harm} & \small\texttt{self-\newline harm} & \small\texttt{sexual} & \small\texttt{toxic} & \small\texttt{violent}\\
\midrule

    & RF & 0.378 & 0.341 & 0.280 & 0.354 & 0.371 & 0.355 & 0.668 & 0.375 \\
    \multirow{-2}{2.3cm}{BERT Large (326m)} & LightGBM & 0.452 & 0.419 & 0.363 & 0.427 & 0.441 & 0.438 & 0.740 & 0.450 \\ 
    \hline

    & RF & 0.360 & 0.328 & 0.273 & 0.327 & 0.350 & 0.329 & 0.612 & 0.363 \\
    \multirow{-2}{3 cm}{BERT Large finetuned (326m)} & LightGBM & 0.433 & 0.410 & 0.352 & 0.415 & 0.405 & 0.399 & 0.713 & 0.433 \\ 
    \hline

\end{tabular}
\caption{Comparison of PR-AUC between combinations of the BGE, both the off-the-shelf and finetuned, and the Random Forest ("RF") and LightGBM classifiers. They are both not as accurate as LionGuard's chosen combination.}
\label{tab:additional_table}
\end{table*}

\clearpage
\section{Examples of predictions on Singlish texts}
\label{sec:appendix-example-predictions}

See Table \ref{table:example-predictions} for examples of predictions on Singlish texts from LionGuard, Moderation API, Perspective API and LlamaGuard. The categories are with reference to our safety risk categories. Ticks represent the model correctly predicted the text as fulfilling the definition of that category, and crosses indicate the model did not detect the text as belonging to that category.

\begin{table*}[h!]
\centering
\setlength{\cmidrulekern}{0.25em} 
\begin{tabular}{m{6cm} m{1.4cm} m{1.4cm} m{1.4cm} m{1.4cm} m{1.4cm}}
\toprule
  \small\textbf{Text} & \small\textbf{Category} & \small\textbf{LionGuard} & \small\textbf{Moderation} & \small\textbf{Perspective} & \small\textbf{LlamaGuard}\\
  \cmidrule(lr){1-1}
  \cmidrule(lr){2-2}
  \cmidrule(lr){3-3}
  \cmidrule(lr){4-4}
  \cmidrule(lr){5-5}
  \cmidrule(lr){6-6}

    \textit{"BAHAHAHA you CECA dogs do love creating alts to reply in an attempt to mend your hurt egos after getting a trashing down by yours truly."} & \texttt{hateful} & \cmark & \xmark & \xmark & \xmark \\ \hline

  \textit{"Many PRs especially Malaysians rejected citizenship, these are the leeches."} & \texttt{hateful} & \xmark & \cmark & \xmark & \cmark\\ \hline

  \textit{"im a bbfa fcp, i got kkj!! you come nearer, i pew pew your pigu!"} & \texttt{sexual} & \cmark & \xmark & - & \xmark \\ \hline
  \textit{"The users here will probably go home and wank to onlyfans girl to get post nut clarity so that they can shit post about the election outcome"} & \texttt{sexual} & \xmark & \cmark & - & \cmark\\ \hline

  \textit{"AMDK whoreshipping starts yang at some sec skools. Gotch AM singers get invited to sing in skools."} & \texttt{toxic} & \cmark & - & \xmark & - \\ \hline
  \textit{"That calls for a celebration with one scum less in this world."} & \texttt{toxic} & \xmark & - & \cmark & -\\ \hline

\bottomrule
\end{tabular}
\caption{Examples of Singlish texts and whether LionGuard, Moderation API, Perspective API and LlamaGuard correctly detected the category.}
\label{table:example-predictions}
\end{table*}

\end{document}